# *Imagination* as Holographic Processor for Text Animation

**Vadim Astakhov**
**Tamara Astakhova**
University of California San Diego
9500 Gilman Drive
San Diego, CA 92093-0715 USA
+1 858 525 5907
astakhov@ncmir.ucsd.edu

**Brian Sanders**
University of California San Diego
9500 Gilman Drive
San Diego, CA 92093-0715 USA
+1 858 822 0707
bsanders@ncmir.ucsd.edu

**ABSTRACT**
Imagination is the critical point in developing of realistic artificial intelligence (AI) systems. One way to approach imagination would be simulation of its properties and operations. We developed two models "Brain Network Hierarchy of Languages", "Semantical Holographic Calculus" and simulation system ScriptWriter that emulate the process of imagination through an automatic animation of English texts. The purpose of this paper is to demonstrate the model and present "ScriptWriter" system *http://nvo.sdsc.edu/NVO/JCSG/get_SRB_mime_file2.cgi//home/tamara.sdsc/test/demo.zip?F=/home/tamara.sdsc/test/demo.zip&M=application/x-gtar* for simulation of the imagination.

## INTRODUCTION: ARTIFICIAL INTELLIGENCE (AI) AND PROCESS OF IMAGINATION

Humans are exceptionally adept at integrating different perceptual signals to create new emergent structures, which results in new ways of thinking. Even in the absence of external stimulus, the brain can produce imaginative stimuli. Some of these imaginative stimuli are dreams and imaginative stories. The imaginative process is always at work in even the simplest construction of meaning, a concept that philosophers Gilles Fauconnier and Mark Turner call "*two-sided blending*" [1]. It is not hard to realize that the imagination is always at work in the subconscious. Consciousness usually views only a portion of what the mind is doing. Most specialists in various areas have impressive knowledge, but are also unaware of how they are thinking. And even though they are experts, they will not reach justifiable conclusions through introspection.

This led us to a conclusion that *Imagination* is the crucial subconscious feature of the creative human mind and that it might provide a basis for other mental functions. We should address this issue with respect to the problem of creating artificial intelligence. It is reasonable to state that a strong artificial intelligence system competitive to the creative human brain should exhibit a certain level of complexity. Rephrasing Searle [2] that artificial hearts do not have to be made of muscle tissue, but whatever physical substance they are made of should have a causal complexity at least equal to actual heart tissue where the term "causal complexity" is reflecting the quantity of causal relations and their hierarchy. The same might be true for an "artificial brain" that might cause creativity and consciousness though it is made of something totally different than neurons, if the "artificial brain" structures share the level of causal complexity found in brains. It's not like building a "perpetum mobile" as some people refer to when attempting to build AI. We do not observe a "perpetum mobile" anywhere, though we can observe consciousness not just in humans. This observation can be made in other highly developed animals [3].

We think that the first step in making an artificial intelligence system conscious would be a mutual simulation of high-level human cognitive functions, such as memory and imagination with large-scale neural networks. Such a system will provide mapping between mental functions, combinations of the firing rate of the neurons, and the specific neuronal architecture; or even some biochemical features of the neuronal structures as suggested by Crick [4]. The "Universal grammar" optimal theory [5] and A Theory of Cerebral Cortex [6] demonstrate how discrete symbol structures can emerge from continuous dynamic systems such as neural networks. Those symbols can be represented as dynamic states over a set of distributed neurons where various symbols can be represented by various states in the same or different neural net. In our model of Hierarchy of Brain Network Languages, we proposed that a neural network's dynamics produce a hierarchy of communication "languages," starting from a simple signal level language, and advancing to levels where neural networks communicate by firing complex nested structures. Such communication is complex enough for syntactic structures to emerge from an optimization of neural network dynamics. We also provide a "Holographic Calculus" as a candidate for neural imprint computing that can lead to the emergence of semantic relations.

Before digging into our model, we would like to clarify that it is essential to distinguish between "primary consciousness," which means simple sensations, and perceptual experiences and higher order consciousness, which include self-consciousness and language. We assume that the higher order consciousness is built up out of processes that are already conscious, that have primary consciousness. In order to have primary consciousness, an AI should possess some mechanisms provided by the human brain. Let's go through the list of features that should be implemented in an AI as a base.

Probably one of the most basic components is memory. The human brain is not just a passive process of storing memories, but is also an active process of re-categorizing on the basis of previous categorizations. Adaptive Resonance Theory [7] one of the candidates is to provide a model of human memory. We modified ART and constructed a neural network that can store and categorize perceived images every time they are perceived rather then comparing them with stored templates. For example, if a child sees a cat, it acquires the cat category through the experience of seeing a cat and organizing its experience by way of the recurrent network [Shennon 8] first introduced by Shennon in 1948 and recently known in neuroscience by name reentrant maps [Edelman 9]. Then the next time the child sees a cat, the child has a similar perceptual input. He or she re-categorizes the input by enhancing the previously established categorization. The brain does this by changes in the population of synapses in the global mapping. It does not recall a stereotype but continually reinvents the category of cats. This concept of memory provides an alternative to the traditional idea of memory as a storehouse of knowledge and experience, and of remembering as a process of retrieval from the storehouse. It also explains the latest claims from recent psychological publications [10] why most of our memories of past events are constructed and have just a few correctly memorized elements. Based on re-categorization, we can easily see how most of our memories can be constructed due to memory re-invention of each category from the most recent perceptual inputs.

Another critical component is the ability of the system to learn. The AI system has to prefer some things to others in order to learn. Learning is a matter of changes in behavior that are based on categorizations governed by positive and negative values.

The system also needs the ability to discriminate the self from the non-self. This is not yet self-consciousness, because it can be done without a discrete concept of the self. The system must be able to discriminate itself from the world. The apparatus for this distinction should provide the "body" system a set of spatial and temporal constraints, and should also register the system's internal states and discriminate them from those that take in signals from the external world, such as that feeling hunger is part of the "self," and the visual system, which enables us to see objects around us.

The AI needs a system for categorizing sequential events in time and for forming concepts. Not only should the AI be able to categorize cat and dogs, but it should also be able to categorize the sequence of events as a sequence. An example of this would be a cat followed by a dog. And it must be able to form pre-linguistic concepts corresponding to these categories.

A special kind of memory is needed to mutually configure interactions among various systems. An example would be, the experience of sunshine for warmth and the experience of snow for cold. The system should have categories corresponding to the sequences of events that cause warmth, or conversely, cold. And its memories are related to ongoing perceptual categorizations in real time.

We need a set of reentrant connections between the special memory system and the anatomical systems, which are dedicated to perceptual categorizations. It's the functioning of these reentrant connections that give us the sufficient conditions for the appearance of primary consciousness.

Using all of these features, we can define primary consciousness as an outcome of a recursively comparative memory, in which the previous self and non-self categorizations are continually related to present perceptual categorizations and their short-term succession, before such categorizations have become part of that memory.

On the other hand, higher-level mental functions should provide:

Conceptual integration is at the heart of imagination. It connects input spaces, projects selectively to a blended space, and develops emergent structures through composition, completion, and elaboration in the blend.

Emergent structures arise in the blends that are not copied directly from any input. They are generated in three ways: through projections composed from the inputs, through completion based on independently recruited frames and scenarios, and through elaboration.

Composition – blending can compose elements from the input spaces in order to provide relations that do not exist in the separate inputs.

Completion – we rarely realize the extent of background knowledge and structure we bring into a blend unconsciously.

Elaboration – we elaborate blends by treating them as simulations and running them imaginatively, according to the principles that have been established for the blend.

Another big area of human behavior involved in development of imagination is human internal "beliefs".

The statement "person X believes Y" is equivalent to a series of conditional statements that assess how X would behave under certain circumstances. In that sense, beliefs are unobservable entities that cause observable traits in human behavior. These unobservable entities originated from facts of observation, from memory, from self-knowledge, and from experimentation. All of these blends in imagination lead to "a belief" through individual inference rules. Imagination is the result of merging and blending various concepts.

Following this list of high-level functions, we analyzed and implemented some aspects of imagination such as integration and identity in the software. We see integration as finding identities, and oppositions as parts of a more complicated process, which has elaborate conceptual properties that can be both structural and dynamic. It typically goes entirely unnoticed since it works so fast in the backstage of cognition. On the other hand the identity is the recognition of identity and equivalence that can be mathematically represented as A=A. It is a spectacular product of complex, imaginative and non-consciousness work. Identity and non-identity, equivalence and differences are apprehensible in consciousness and provide a natural beginning place for formal simulation approach. Identity and opposition are final products provided to consciousness after elaborate unconsciousness work and they are not a primitive starting point.

These operations are very complex and mostly unconscious for humans but at the same time play a basic role in the emergence of meaning and consciousness. From everyday experiences of meaning and human creativity, we can conclude that the meaning and basic consciousness operation lies in the complex emergent dynamics triggered in the imaginative mind. It seems reasonable to imply that consciousness and mind prompt for massive imaginative integration.

We believe that simulation of imagination is a first step for building a powerful AI system. To accomplish that step, a pluggable architecture called "ScriptWriter" was developed. That provides us with the ability to simulate imagination through the process of text animation.

## EMERGENCE OF HOLOGRAPHIC NETWORK PROCESSOR AND HOLOGRAPHIC CALCULUS

A holographic network is organized as a graph shaped hierarchy of nodes, where each node is a network itself and implements a common learning and memory function. *Figure 1* represents a process of development for the new holographic network.

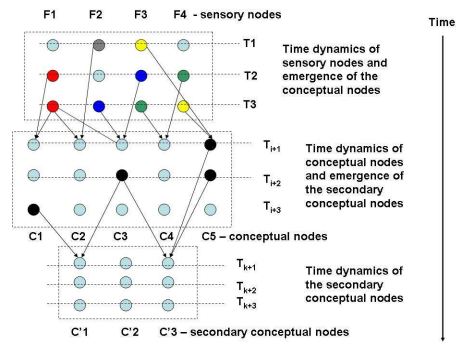

*Figure 1 Emergence of the holographic network represented in time as a process of development hierarchy of conceptual nodes through assembly of temporal signals in patterns called "concepts. Those concept it-self can participate in the process or development of new nodes.*

First level of holographic network is a set of sensory nodes. In a human, the optic nerve that carries information from the retina to the cortex consists of about one million fibers where each fiber carries information about light in small part of visible space. The auditory nerve is about thirty thousand fibers, where each fiber carries information about sound in a small frequency range. The sensory node is like a fiber where each input noted measures a local and simple quantity.

Co-occurrence of signals on different sensory nodes leads to emergence of a so called primary conceptual node that is implemented as a coherent firing of some sub-net of neurons. Same idea applied to the sequence of signals firing within certain time window. The sequence of firing (T1, T2, and T3) or co-firing among sensory nodes will lead to emergence of a primary concept node as an assembly of those signals. In such a view, the primary concept is an internal representation of a spatial and temporal activity pattern of sensory nodes. The assembled pattern can be represented as a dynamic state of an underlying neural network.

Repeating of the same pattern with minor variations will increase strength of underlying neural connections and enhance strength of introduced concepts. By repeating the same pattern, the assembled concept node is getting *re-called* and updated. It stores the signature of the current pattern as well as previously observed. This way the concept keeps a history of the pattern evolution.

If variations of incoming new sensory patterns are sufficiently large then the new primary concept introduced. Those primary concepts can interact with each other and sensory concepts due to interactions among underplaying neural networks.

That network is a Bayesian network in which inferences represented by edges emerge as probability of nodes co-occurrence (be updated –accessed from sensory nodes) within certain time window. We propose the formalism of those interactions through circular convolution and de-convolution which are eventually analogies of holography in optics. For any two random vectors X and Y, the circular convolution will produce another vector

$$z = x \mathbin{@} y, \text{ where } z_j = \sum_k x_k * y_{j-k}$$

Convolution/de-convolution propagates as diffusion through network of concepts. Thus nodes interact (convolve/de-convolve) "holographically" with each other and can produce new imaginary nodes. Same as for sensory nodes, the primary concepts will create a new layer of secondary concepts through the same mechanism of repeating sequences and co-occurrence within some time-window. Those secondary, primary and sensory nodes also can interact and re-currently lead to emergence of higher conceptual levels. Those interactions obviously create inter-connections among nodes from various levels.

To not over extend a memory and keep the only significant experience, the intensity of each stored pattern signature exponentially decays. Each previously recorded pattern has a decay time dependant on the amount of secondary patterns emerged through co-occurrence with other concept during the time when the pattern was experienced.

*Figure 2 Patterns recorded at different period of time represented by their signatures decaying in time. Figure represent reinforcement of the signature recorded at moment T(k+2) for concept C3. Reinforcement performed due to C3 occurrence in emergence of new secondary conceptual nodes (Figure 1) through assembly with C1 and C5 (Figure 1)*

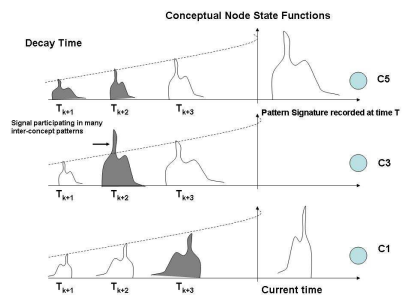

Figure 2 illustrate how concept "C2" can stay longer in memory due to its participation in the emergence of several secondary concepts " C'1 and C'2 " through assembly with primary concepts "C5 and C1" respectively. Figure 2 demonstrate *signatures* of concepts decaying in time. Those signatures are a numerical representation of dynamic states of underlying neural networks. If one concept participate in many assemblies which lead to the emergence of the new secondary nodes, then the concept gets reinforced each time the new secondary node created or re-called.

If the node does not participate in the assembly of any other nodes, then it get decayed without reinforcement. The concept node keeps a signature as a vector in which coordinates represent recorded signals from sensory nodes multiplied by decay time exponent: $S1*\exp(-t/d)$. The decay time "d" will be represented as the amount of inter-node connections that will maximize the probability of the node to be re-enforced due-to various interactions.

When intensity of the stored signature becomes less then some pre-defined threshold, then the signature vanishes from the holographic network. If all signatures recorded at different time for the concept vanish then the concept vanishes. This way the holographic network keeps itself adapted to current experience and eliminates old experience (signatures) and even old concepts. It also eliminates "conceptual noise" that is a bunch of new patterns that was stored and lead to the emergence of new primary concepts. Those concepts do not get any further support through repeating re-occurrences of their patterns. At the same as we demonstrate on *figure 2* the pattern participating in assembly of new concept nodes can be often reinforced by its assembly members and thus stay in "memory" even after long period of non-recall.

We introduce conceptual node signature to represent dynamic states of the underlying network. Figure 2 illustrates a signature as a one dimensional vector but it is actually multi-dimensional complex (p-adic to keep order) vector. Each elementary signal from sensory network is coded by set of neurons with different phases. Figure 3 gives an imaginary analogy where each pixel of 2-D image can be represented as a concentric curve of another 2D image. We call such transformation as *delocalization*. That transforms all local features of the original image to distributed representation. It is very similar to the process of holography recording in optics. We call the new image as *holographic map*.

*Figure 3 show process delocalization through projection of each point of original image to a concentric curve of the new image.*

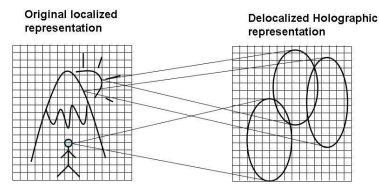

Each pixel on the new distributed representation of the original image can be seen as a neuron that keeps information about original pixel and has some internal phase information that distinguishes its state from states of other neurons. Thus each sensory vector (matrix) has a delocalized holographic representation on the neural matrix that keeps local sensory information as a set of intensities and phases of distributed neurons. Such representations let us realize circular convolution on neural networks and implement holographic calculus.

*Figure 4A show inter-connection among neurons forming holographic representation of sensor signals as well as connections among neurons from different holographic representations (maps). Re-entrant connections among holographic maps provide a mechanism for holographic convolution/de-convolution calculus. Figure 4B show inter-connection among neurons involved in holographic representation of a local signal.*

"A"     "B"

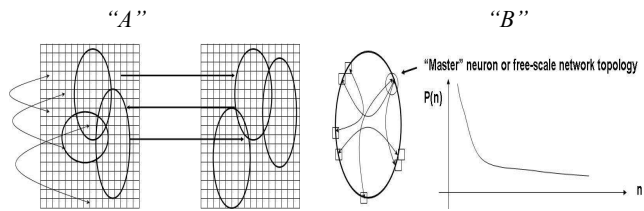

Inter-connection among neural networks provides a mechanism for concept reinforcement from other concepts. We introduce inter-connections among those neurons with scale free-distribution (Figure 4B). Such architecture of the network will produce limited amount of "muster" nodes connected to all other. That topology will provide a light mechanism to re-call or reinforce the network through accessing only master neurons which will propagate the assessing signals to all others.

Figure 5 illustrates the process of reinforcement of some features of the old memories through convolution with new experience. Holographic representation of stored signatures that we called "holographic map" can be seen as some kind of "inner image". Categorization emerges through the reinforcements of different features of stored images. Such reinforcement is result of assembly with other concepts or re-calls (update) from sensory nodes.

*Figure 5 shows image/concept "A" stored in memory and reinforced by new experience "B"*

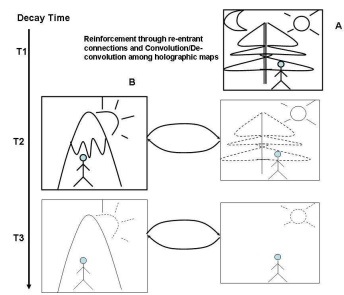

Convolution and de-convolution provide nice mechanisms for concepts and image operations. Any activity in a sensory or conceptual node will propagate through network of conceptions due to underlying network connections. Such propagation described by circular convolution and de-convolution is an analog of a holographic process. That activity of certain nodes will be a result of convolutions that lead to emergence of a new set of nodes as a holographic reconstruction.

Propagation through those reconstructed nodes will lead to chain holographic reconstructions even further. This process will create waves of holographic reconstruction that are never ended and affected by external sensory node stimulation. Such holographic propagation leads to emergence of *prototypes* such as "tree" and "human"

*Figure 6 shows convolution/de-convolution among of initial 2D-sensory signals into "internal concepts". Further convolution/de-convolution of concept signals into prototypes.*

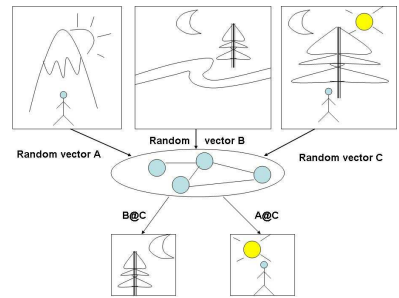

Inner concepts continue process of further convolution/de-convolution among emerged concepts. Those concepts that will get higher re-informant through assembly with others or re-call from sensory nodes will survive and lead to emergence of prototypes.

## ONTOLOGICAL MODEL FOR BLENDING PROCESS OF IMAGINATION

### Concepts

We proposed "Topological whole image computing" which based on whole image transformation and segmentation rather than starting from some elementary or primitive objects. Computing based on primitive concepts can be represented as symbolic computing but this "topological" approach deals with the non-local representation of a whole scene within the neural network of the AI-brain and integrated transformations over such representation. These transformations are neural networks, where primitive objects emerge as local invariants within scene. We assume that, yet there is no supreme area or executive program binding the color, edge, form, and movement of an object into a coherent percept.

Objects represented as a whole in the juvenile AI-brain, as well as features such as color and shape, will diverge later over brain development.

A coherent perception in fact nevertheless emerges in various contexts, and explaining how this occurs constitutes the so-called binding problem.

The behavior of human infants conveys signs of strong synesthesia. So, we are suggesting that there is no "binding" in the juvenile brain, because it has not developed to the point where it can break perceptional fields into a set of modalities. For an adult brain the situation is different. A healthy and developed adult brain is well specialized for modalities like color, shape and others.

The "Topological" model takes binding as reentry mapping between distributed multi-model imprints from early childhood and features of a perceived field.

We propose to simulate mutual reentrant interactions among our holographic neuronal groups. For a time, various and linked neuronal groups in each map (neuronal group specialized for specific dynamics representing imprint, an object or some feature) to those of others to form a functional circuit. The neurons that yield such circuits fire more or less synchronously. (They provide a holographic representation of an object or an object feature.) But the next time, different neurons and neuronal groups may form a structurally different circuit, which nevertheless has the same output. And again, in the succeeding time period, a new circuit is formed using some of the same neurons, as well as completely new ones in from a different group. These different circuits are degenerate, that is, they are different in structure yet they yield similar outputs to solve the binding problem. Such a multiple-implementation can be realized in the holographic model.

As a result of reentry, the properties of synchrony and coherency allow more than one structure to give a similar output. As long as such degenerate operations occur in the correct sequence to link distributed populations of neuronal groups, there is no need for an executive program as there would be in a computer.

To construct an algorithm and a dynamic system, which emulates mental functions, fundamental theoretical units must be chosen. We propose the term "concept" for sub-net of neurons exposing certain dynamical properties and at the same time is a internal representation of an object occupying space and time, an object with attributes specifying what an object is or does and what relations exist between objects. Example: concepts "woman", "walk", "beach" can lead to the conceptualization "A woman walked on the beach" that will lead to an animated image of a woman walking on the ocean beach. This conceptualization will imply many "beliefs". One possible belief here can be - "the woman wears something". Many of us intuitively "believe" that people usually wear something when they "walk" if the opposite is not mentioned. That "belief" will cause the imagination of many people to provide an image of the woman walking on the beach and wearing clothing, even if nothing was mentioned about clothing. It is a totally different case where we have the sentence: "A naked person is walking on the beach".

The following relevant rules can exist in the system:

IF X is like Y then X seeks Y.

IF Y disturbs X then X avoids Y

But sometimes such rules can be in a conflict that leads to the emergence of new, blending structures.

**Universal Structure of objects in the scene**

A participant in the scene entity is assumed to be in one of the three states (Active Actor, Passive Actor and Action) with a binding pattern for every disclosed relation. Every entity's element (relations and attributes) has a descriptor for a keyword search, and a so-called semantic-type that can be used to map the element to its ontology. For example: Relation – behind (far behind) has a type-position/orientation. Another example: Attribute – red has type-color.

Further, an entity may disclose a set of functions that are internally treated as relations with the binding pattern (b, f), where b represents a set of bound arguments and the single f is the free output variable of the function. For example, "take the ball" can be treated as a human specialized function, which is used to raise the human actor hand in a set of specified scenes. Such functions will depend on sets of binding parameters "b" that they characterize, or the position of the ball and return "f"-position of the hand. Such a model lets us treat animation as Petri Net dynamics with computations where actors-nodes take different states in time.

**Mental Space**

We use a term "mental space" as small packets of concepts, which are constructed as we think and talk. Also, we have *Conceptual integration* as a critical part of imagination. It connects input *mental spaces*, projects selectively to a blended imaginary space, and develops emergent structures through composition, competition, and elaboration in the blend.

For example, a set of sentences: "The blue ball was left on the beach. A woman walks on the beach," imply two input concept spaces "Woman walks on the beach" and "ball was left on the beach".

We perform *Cross-Space mapping* which connect counterparts in the input mental spaces and then construct *Generic Space* that maps onto each of the inputs and contains what the inputs have in common: beach, ocean, and horizon. The final *blending* does the projection of the ocean beach from the two input mental spaces to the same single beach in blended imagination space.

Such blending develops emergent imaginative structures that are not present in the inputs like "woman walk toward the ball" or "woman walk relatively close to the ball". It seems intuitively obvious that imagination can create an integrated scene with all the mentioned objects as a result of those two sentences.

**Ontologies**

Ontology is a term-graph whose nodes represent terms from a domain-specific vocabulary, and whose edges represent relations that also come from an interpreted vocabulary. The nodes and edges are typed according to a simple, commonly agreed upon set of types produced by test-bed scientists. The most common interpretation is given by rules such as the transitivity of *is-a* or *has-a* relations, which can be used to implement inheritance and composition. However, there are also domain specific rules for relationships such as *region-subpart* (rock-region -> mountain-region) and *expressed-by* (emotion-state -> face) that need special rules of inference. For example, if a rock-region participates in an imaginary scene (such as "he climbs the rock") and the human-emotion is expressed-by a face, then the rocks case is an emotion that will be expressed on the face.

In the current *ScriptWriter* framework, ontologies are represented as a set of relations comprised of a set of nodes and a set of edges, with appropriate attributes for each node and edge. Other operations, including graph functions such as path and descendant finding, and inference functions like finding transitive edges are implemented in Java.

We build ontology by extracting relations pair-wise between English words like "head –part of -body". We also assign a wait for each relation that reflects the probability to have two words in one sentence or in two concurrent sentences. That probability was extracted as a frequency of pair-wise occurrences of the two specified words. To perform a calculation, a test cohort of the fiction texts was collected.

*Ontological dK-series and dK-graphs*
Ontological graphs created the way described above are dependent on the cohort of the text and cannot pretend to be generic enough. Also, even for a small dictionary of English words this is extremely complex. Here, we need a way to approximate properties of a generic ontological graph, that can be built on a limited text cohort but that can capture topological properties of generic English text.

To capture such complexity of graph properties, we use the dK-graphs approach [6]. This approach demonstrates that properties of almost any complex graph can be approximated by the random graph built by set of dK graphs :0K, 1K, 2K and 3K where "K" is the notation for a node degree and d-for joint degree distribution that d node of degree "k" are connected.

Based on our text cohort, we estimate 0K-average node degrees as the average frequency of a word, 1K –node degree distribution as frequencies for the words in the text cohort, 2K and 3K – joint degree distribution were extracted as pair-wise and triple-wise frequencies of having two/three words in two/three sub-sequent sentences. Those values were assigned for each dK graph of our ontology and later used during operation of *ontological confabulation*.

*Term-Object-Map*
We have a specific source called the *term-object-map* that maintains a mapping between ontological terms and 3D-animation objects library, which was developed for the Maya animation environment. These objects are used by the system to build animation.

*Mapping Relations*
Currently in the animation industry, the burden of creating complex animated scenes over many actors is placed on the animation specialist, who works hard to capture the requirements of the script at hand. This leads to the pragmatic problem that the relationships between attributes disclosed by different objects and between object parameters are, quite often, not obvious. To account for this, the system has created additional *mapping relations*. Currently there are three kinds of mapping relations.

The *ontology-map* relation that maps data values from an object to a term of the ontology

A joinable relation that links attributes from different objects if their attribute types, relations and semantic types match.

The *value-map* relation which maps a fuzzy parameter value (speed fast) to the equivalent attribute value disclosed by the animation software.

## UNIVERSAL TEXT FILTERING AND dK-ONTOLOGY CONFABULATION
### Text filtering
ScriptWriter uses simple English text as an input and generates output animation. First, it performs text processing to extract semantic relations among words in the sentences. Mental space is created for each sentence. As an example, we consider the simple text of three sentences: *"A woman walks on the beach. The blue ball was left on the beach. A woman takes this ball"*.

Fig.9 represent details of the first mental space that are built from the sentence "Woman walks on the beach". The sentence was processed and its *Universal structure* was extracted: "Active actor (woman)–action (walk)-passive actor (beach)". Instances of the universal structure (woman, walk and beach) were anchored (colored by yellow) to an ontological graph that represent relations among concepts. We perform graph expansion operations to integrate all relevant objects required for the mental space such as ocean, sky and the woman's clothing, which are not mentioned in the sentence (colored by red).

Each concept such as "beach", "woman", and "ball" represents a sub-graph that connect all concepts relevant to the specified term.

*Figure 7. Part of ontological graph that represent mental space for sentence "Woman walks on the beach"*

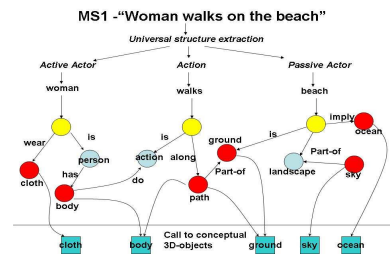

Same operations were performed for the second and third sentences.

## Graph calculus and Generic Space

Generic space was constructed by the mapping of objects such as "woman", "beach" and "ball", which co-occurred in different *mental space.*

Starting from those concepts we perform node expansion as a procedure of finding neighbor concepts connected to generic concepts; for example, the "color" of the ball and "body" of the woman. We perform node expansion by using various relations (represented by edges on the ontological graph) such as analogy/disanalogy, cause-effect, representation, identity, part-whole, uniqueness, similarity, and various properties. All these classes of relations are represented by various grammar constrains in the English texts. Follow the ideas of "Universal grammar" and "Distributed Reduced Representation" proposed in papers of Paul Smolensk [8], we define any semantic space as a convolution of some vectors-concepts providing the space reduced representations. Rather then using Smolensky' Tensor [7] that has the variable length we decided to use the Holographic reduced representation techniques [8]

Each node and edge were assigned to a random vector from 512-dim space then any combination of connected nodes and edges were defined as a result of circular convolution on the vectors [8]: $z=x @ y$, where $z_j = \sum x_k * y_{j-k}$

And indexes represent coordinates in 512 dimensional space. Such representation will provide coding schema to any complex scene.

*Example: sub-graph "woman-wear-clothing" was represented as $z=x @ y = x @ ( m @ n)$; where "m"- represents vector "woman", "n"-"wear", "clothing" and convolution of "m" and "n" gave us "woman-wear-" open-end sub-graph that has one node and one edge.*

## Renormalization as dimension reduction

Due to the high complexity of the ontological graph we performed a "graph compression" that resulted in elimination of some redundant links. We decided to perform compressions just over the vital relations such as Time, Space, Identity, Role, Cause-Effect, Change, Intentionality, Representations and Attributes. Mathematically, that operation was implemented as circular correlation: $y=x\#z$, where $y_j = \sum x_k * z_{k+j}$. Taking the previous example, that operation is equivalent to

"woman-wear-"#"woman-wear-clothing"="clothing" and will return the node "clothing".

Holographic reduced representation let us quickly compute "generic space".

The final ontological sub-graphs blending and generation of resulting imaged space were performed as a confabulation on the *generic space*.

## dK- Ontological Confabulation

We extended the operation of Confabulation previously proposed in paper [7] and developed an extended version of this operation for ontological graphs. This operation extracts concepts not mentioned in the text message. Consider our example: "A blue ball was on the beach. A woman walks on the beach. She takes the ball and kicks it". It seems clear that our imagination should build the picture of the woman that binds her body to take the ball by hands, even though nothing in the text mentioned that bio-mechanical process. To do that, *Imaginizer* will use *ontological confabulation* for extracting knowledge associated with provided concepts. Confabulation was defined [9] as a maximization of probability to start from nodes a, b and c and get node d: $p(abc|d) \sim p(a|d) * p(b|d) * p(c|d)$

We start from any input node A, and then randomly walk and calculate its probability to get to node B. That probability obviously depends on the order degree for node B. The more nodes connected to B through some path, the higher probability it is to get there. The initial algorithm proposed in [9] uses only the weight of the edges that were calculated from pairwise frequencies of two nodes in some text.

Here we propose a new algorithm to calculate the probability of transitions by using node degrees and joint probabilities of dK –series (0K, 1K, 2K, 3K) that was extracted from the cohort of the text during building of the ontological graph. Due to analysis [6] that any properties of the complex graph can be reconstructed by a random graph with identical statistical properties for 0 to 3K sub-graphs, we suggest calculating the probability of a transition from A to B through some intermediate nodes as a sum over degree distributions for all intermediate 0K,1K,2K and 3K sub-graph between A and B.

If we start from several input nodes, then the total probability to get to the node B is sum over all probabilities calculated for each input node. The B node with highest probability will be taken as a part of new blended space. The next less probable node was taken as a part of the imagined space if its probability was higher then some threshold. That threshold was estimated heuristically.

*Figure 8.. Final imaginary space emerges as a result of ontological confabulation.*

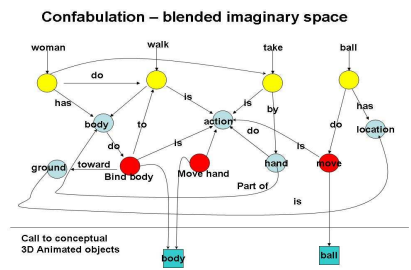

Each scene represented as a vector as well as graph of relevant concepts. Dynamics of underlying neural groups can implements various operations on graphs. Those operations can be illustrated using holographic analogy from optics. Figure 6 illustrate de-convolution of the Blended space with two vectors (Sub-scene 1 and Sub-Scene 2) which represent various sub-scenes. The result vectors called "Holography" due to similarity between mathematical formalism for our graph operators and optical holography. The result of de-convolution is holographic reduced representation for some portion of our blended graph. Those portions obviously depend on the de-convolution vectors. That illustrated on figure 7 where each resulting vector encodes a graph of concepts related with each. Encoded graphs are internal representation of visual scenes in the "ScriptWriter brain".

*Figure 9. Part of ontological graph that represent mental space for sentence "Woman takes the ball"*

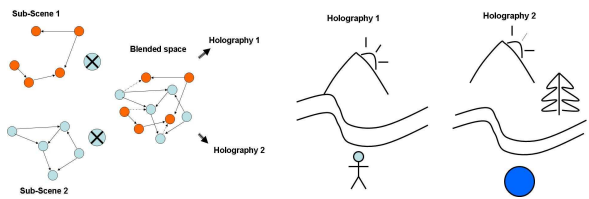

## "SCRIPT WRITER" AND ANIMATION PROCESS

*ScriptWriter* is an attempt to simulate visual imagination through the creation of 3-D animation of English text. It is done in such as way to approximate a human reading a story and imagining it from a first person perspective.

The first limited version of the software called *ScriptWriter* we built provides us with the ability to "imagine" short stories with primitive objects such as humans, actors, and several landscapes. *ScriptWriter* performs text processing, semantic extraction and animation planning that utilizes approaches tested in the various areas of Interactive Virtual Environment development [2,3,4,5]. Those were designed specifically for the development of believable agents – characters which express rich personality, and which, in our case, play roles in an imaged animated world. ScriptWritet platform provides a set of libraries and APIs. We propose a Concept Mapping tool which integrates ontology with mapped objects stored in a relational database. The ontology is described using OWL (Ontology Web Language) which is built on top of RDF (Resource Description Framework), which can be edited using a tool such as Protégé. An OWL ontology is constructed with a hierarchical vocabulary of terms to describe concepts in blended space. Each concept in the ontology maps to a Data Object, which is a set of fields and values stored in a database. There is also the capability to retrieve related data fields via sequences of foreign key/primary keys and map those field values. The full set of concept-data object mappings is saved to a project which can be loaded into a database. It is then possible to perform a join query using concept terms to retrieve all relevant data stored in the database as a result set. This method allows ontological concepts to be used as a generalized query vocabulary for database information retrieval.

It provides a connection to sensory-motor system of agents – "Actors" and support multi-agent coordination. *ScriptWriter* scenario was organized as a collection of the behavioral actions of the actors or simply "Actions". During the imagination process some of the images and their actions encapsulate some other actions in the same kind of nested way that will produce sequential behavior. An example of sequential behavior is shown below: "A woman walks on the beach. There was a blue ball on the beach. She kicks the ball." We easily can imagine a scene where a woman walks to the ball left on the beach by someone and how she is kicking the ball. The text can be animated now by *ScriptWriter* with minimal human intervention.

## THE DEMONSTRATION AND FUTHER DIRECTIONS

The demonstration will present to the user the "ScriptWriter" performing animation for short texts of 2-4 sentences. This will include a demonstration of text processing and building semantic relations as well as a generation of a scenario for the animation. This scenario will be used to automatically generate a script for building animation in the "Maya" animation environment.

*Figure 10. ScriptWrite screen shot*

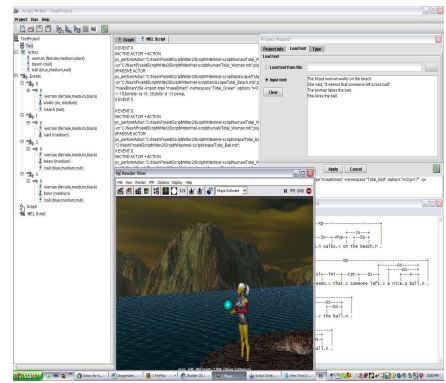

This project raises a number of interesting AI research issues, including imagination process management for coordinating visual object interactivity, natural language understanding, and autonomous agents (objects, landscapes and their interactions) in the context of a story. These issues were partially answered in the current version and will be refined in the next generations of the software.

We also suggest simulation of the new network processor architectures based on proposed holographic calculus.

## ACKNOWLEDGMENTS
We thank Edward Ross and David Little from University of California San Diego for discussion and comments.